\documentclass[lettersize,journal]{IEEEtran}

\usepackage{amsmath,amssymb,amsfonts}

\usepackage{algorithm}
\usepackage{algorithmic}

\usepackage{graphicx}
\usepackage[caption=false,font=normalsize,labelfont=sf,textfont=sf]{subfig}
\usepackage{float}
\usepackage{enumitem}

\usepackage{booktabs, multirow} 
\usepackage[table]{xcolor}     
\usepackage{textcomp}
\usepackage{soul}               
\usepackage{url}
\usepackage{cite}

\usepackage{stfloats}    
\usepackage{changepage,threeparttable}

\usepackage{hyperref}

\let\OLDthebibliography\thebibliography
\renewcommand\thebibliography[1]{
  \OLDthebibliography{#1}
  \setlength{\parskip}{0pt}
  \setlength{\itemsep}{0pt plus 0.3ex}
}

\newcommand{\etal}{\textit{et al.}}

\begin{document}

\title{Detecting Lip-Syncing Deepfakes: Vision Temporal Transformer for Analyzing Mouth Inconsistencies}

\author{\IEEEauthorblockN{Soumyya Kanti Datta, Shan Jia, Siwei Lyu } \\
\IEEEauthorblockA{\textit{University at Buffalo, State University of New York} \\
\{soumyyak, shanjia, siweilyu\}@buffalo.edu}

}




\maketitle

\begin{abstract}

Deepfakes are AI-generated media in which the original content is digitally altered to create convincing but manipulated images, videos, or audio. Among the various types of deepfakes, lip-syncing deepfakes are one of the most challenging deepfakes to detect. In these videos, a person's lip movements are synthesized to match altered or entirely new audio using AI models. Therefore, unlike other types of deepfakes, the artifacts in lip-syncing deepfakes are confined to the mouth region, making them more subtle and, thus harder to discern. In this paper, we propose LIPINC-V2, a novel detection framework that leverages a combination of vision temporal transformer with multihead cross-attention to detect lip-syncing deepfakes by identifying spatiotemporal inconsistencies in the mouth region. These inconsistencies appear across adjacent frames and persist throughout the video. Our model can successfully capture both short-term and long-term variations in mouth movement, enhancing its ability to detect these inconsistencies. Additionally,  we created a new lip-syncing deepfake dataset, LipSyncTIMIT, which was generated using five state-of-the-art lip-syncing models to simulate real-world scenarios. Extensive experiments on our proposed LipSyncTIMIT dataset and two other benchmark deepfake datasets demonstrate that our model achieves state-of-the-art performance. The code and the dataset are available at \url{https://github.com/skrantidatta/LIPINC-V2}.

\end{abstract}

\begin{IEEEkeywords}
 Deepfake detection, Lip-syncing deepfake, Multihead cross-attention, Spatiotemporal inconsistency, Vision Temporal Transformer.
\end{IEEEkeywords}

\section{Introduction}

The rapid advancements in computer graphics and generative AI technologies have revolutionized the creation of synthetic media, enabling the generation of highly realistic deepfakes. Deepfakes are digitally manipulated media, that include fake audio, videos, and images of objects such as vehicles, animals, as well as humans. Among these, deepfakes involving humans, especially those with human faces, are very prominent. These deepfakes are generated by digitally altering a person's appearance, voice, or actions in videos to mimic someone else. Human face deepfakes can be categorized into four types: face swapping~\cite{faceswap}, face reenactment~\cite{Faceforensics++}, lip-syncing~\cite{lipsync}, and face animation~\cite{zeng2022fnevr}. These techniques are powered by advanced generative models, such as  generative adversarial networks (GANs), autoencoders, and diffusion models. The ongoing evolution of these technologies has significantly blurred the line between real and synthetic media, posing challenges for both human perception and automated systems in reliably distinguishing real content from manipulated media.

Video deepfakes, while raising ethical concerns, also offer valuable applications when used responsibly. In entertainment, they can enhance movie dubbing, create personalized virtual avatars, and enable actors to portray different ages. In education, deepfakes can be used to implement interactive learning through historical reenactments and customized tutorials. In healthcare, they can aid medical training with realistic patient simulations and assistive technologies. They can also be used to preserve cultural heritage by recreating historical figures and reviving endangered languages. However, deepfakes come with significant risks, particularly in disinformation, identity theft, and fraud. They enable the creation of highly convincing videos or audio clips in which individuals appear to say or do things they never actually did. Such fabricated content can be weaponized to spread false information, manipulate public opinion, or influence political outcomes. Deepfakes can also harm reputations, disrupt personal and professional relationships, and compromise sensitive data through impersonation or deceit. These manipulations undermine public trust, blur the line between reality and fabrication, and pose challenges to the integrity of digital media, thus requiring advanced tools to detect and mitigate their misuse.

 Recent incidents involving deepfakes have highlighted their potential for misuse, particularly in politics and financial scams. In July 2024, a deepfake video of U.S. Vice President Kamala Harris circulated, depicting her delivering a fabricated speech. The video, reposted by Elon Musk on his platform X (formerly Twitter), drew significant criticism for its potential to mislead the public during an election year. Musk defended the video as satire, but the incident underscored the risks of AI-generated content eroding public trust \cite{ap_parody_ad_ai_harris_musk,ie_elon_musk_kamala_harris_deepfake}.
 Financial scams involving deepfakes have also become increasingly sophisticated.
 In August 2023, a man from Brighton lost £76,000 after falling for a Facebook advertisement featuring AI-generated videos of Martin Lewis and Elon Musk endorsing a fake bitcoin investment scheme\cite{bbc_martin_lewis_deepfake}.
 Similarly, in February 2024, an article was published that stated that  scammers used deepfake technology and AI voice cloning to steal \$25 million by impersonating the company's chief financial officer during a video conference \cite{ChenMagramo2024}.
 In November 2024, another article stated that a UK woman was deceived out of £20,000 by a deepfake video of a U.S. Army colonel, who used AI-generated content to gain her trust and solicit money under false pretenses \cite{thesun_ai_scam}.
 These incidents demonstrate the growing sophistication of deepfake technology and its potential to cause harm in various domains.



The widespread availability of open-source deepfake and AI generated content creation tools has intensified these challenges. Platforms such as Wav2Lip~\cite{lipsync}, Diff2Lip \cite{diff2lip}, DeepFaceLab~\cite{DeepFaceLab-paper}, Ideogram \cite{Ideogram} and FakeApp~\cite{FakeApp} allow users with minimal technical expertise to produce highly realistic deepfake videos. This democratization of deepfake technology has lowered the barrier for malicious actors to misuse these tools, thereby  increasing the demand for precise and efficient detection methods to identify manipulated media.

Lip-syncing is a type of video deepfake in which a person's lip movements are digitally manipulated to synchronize with a specific audio. By seamlessly aligning visual and audio elements, this technique produces a convincing video that shows the person speaking the provided speech. Recent advances in deepfake detection propose to capture the subtle inconsistencies between the video and the audio \cite{agarwal2023watch,oorloff2024avff, feng_et_al}. These approaches analyze correlations between audio and visual signals to detect mismatches. For example, lip movements that fail to align with spoken words create audio-visual dissonance, which can be a key feature  in exposing lip-syncing deepfakes. However, existing methods often rely on motion-based, frame-level, or synchronization-focused features, which may fail to capture all inconsistencies in manipulated videos. For instance, models that rely solely on adjacent frames cannot detect inconsistencies that occur over longer time spans, particularly between non-adjacent frames. On the other hand, models that detect inconsistencies in audio-visual synchronization struggle when the audio is distorted or noisy, or when the manipulation keeps the lip movements and sound relatively synchronized. Such limitations hinders the model's ability to effectively detect complex or partially manipulated lip-syncing deepfakes, leading to false positives or reduced performance. This highlights the need for more robust and comprehensive detection techniques to address the growing sophistication of deepfake technologies.

\begin{figure*}[t]
\centering
\includegraphics[width=\textwidth]{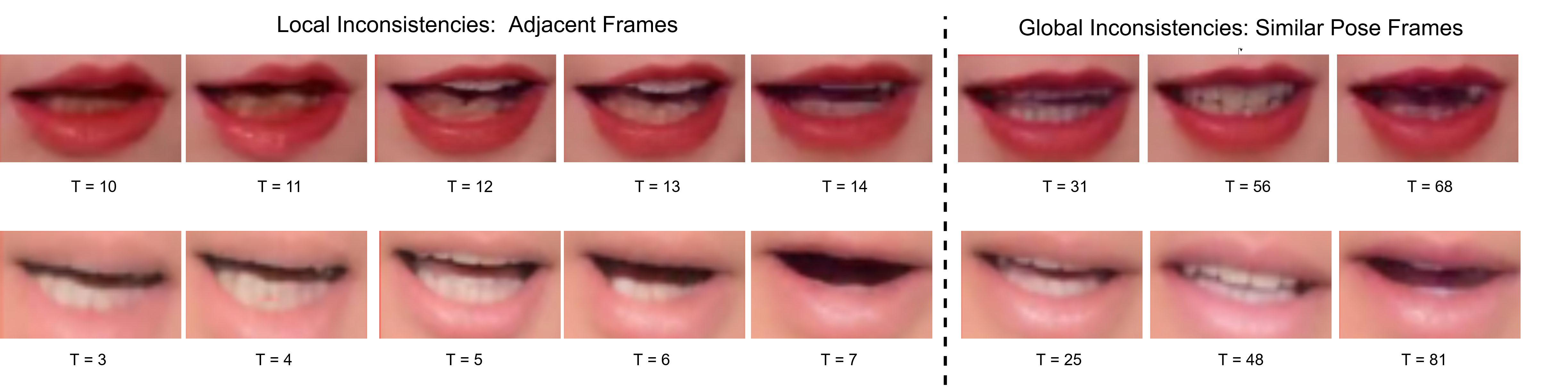}
\caption{\em \small Illustration of the mouth inconsistency in lip-syncing deepfakes. We visualize video frames from two lip-synced videos. Here T represents frame number.
The first five columns present consecutive frames which are 0.03 secs apart for local comparison, while the last three columns offer a broader perspective by displaying frames with similar poses from the entire video, defined as global inconsistencies in our paper. The deepfakes exhibit more pronounced inconsistencies in aspects such as mouth shape, coloration, dental structure, and tongue appearance.} 
\label{fig:example}
\end{figure*}

In this paper, we present a novel model designed to identify spatial-temporal inconsistency patterns to differentiate lip-synced videos from authentic ones. Unlike prior methods that primarily emphasize motion or synchronization, our approach is motivated by two key observations. 
\begin{itemize}
    \item In real videos, mouth movements in successive frames adhere to consistent characteristics, a property that we call as {\it local consistency}. 
    \item Similarly, mouth movements in similar poses across different segments of a video exhibit uniformity, a property that we call as {\it global consistency}.
\end{itemize}
Maintaining such consistency across both local and global contexts proves challenging for lip-synced deepfakes. As demonstrated 
in Fig.~\ref{fig:example}, lip-synced videos often reveal discrepancies in mouth shape, color, and the appearance of teeth and tongue. These spatial and temporal irregularities, while subtle and difficult for the human eye to detect during real-time playback, can be identified through systematic video analysis to uncover patterns of inconsistency.

Our approach identifies inconsistencies using a novel module, the {\it Mouth Spatial-Temporal Inconsistency Extractor} (MSTIE), guided by a specifically designed inconsistency loss function. The MSTIE module processes both spatial and temporal information from the mouth region, extracting fine-grained features critical for distinguishing real videos from deepfakes. By integrating local and global frames, the model captures short-term and long-term variations in mouth movements, enhancing its ability to detect inconsistencies effectively. This dual consistency analysis enables the model to perform comprehensive checks, leveraging localized and temporal context. As a result, our approach achieves state-of-the-art performance in identifying lip-syncing deepfakes.

We conducted extensive experiments on three benchmark datasets 
to evaluate the effectiveness of our approach. The results demonstrate the robustness of our model in detecting both in-domain and cross-domain deepfakes. 


In summary, our work has the following main contributions:
\begin{itemize}[leftmargin=1em, itemsep=0em]
    \item We highlight {spatial-temporal inconsistencies} in the mouth region across both adjacent and non-adjacent frames with similar mouth poses as robust indicators for identifying lip-syncing deepfake videos.
    \item We propose a novel lip-syncing deepfake {detection pipeline}, which includes modules for locally and globally matching mouth poses and a feature extractor for capturing spatial-temporal inconsistencies. The pipeline is enhanced by a unique {inconsistency loss function}, improving the model's detection accuracy.
    \item We introduce the {LipSyncTIMIT dataset}, a collection of lip-syncing deepfakes generated using five state-of-the-art generative models. This dataset is specifically tailored to advance research in lip-syncing deepfake detection and promote the development of more generalizable detection techniques.
    \item Through extensive evaluations on three distinct lip-syncing datasets—{FakeAVCeleb\cite{FakeAVCeleb}}, {LipSyncTIMIT}, and {KODF\cite{Kodf}}—we demonstrate that our method achieves superior performance in detecting both in-domain and previously unseen lip-syncing deepfakes.
\end{itemize}


A preliminary version of this work was presented in our paper \cite{LIPINC}, where we introduced a cross-attention-based model, called LIPINC. This model incorporates a module for locally and globally matching mouth poses, integrating features extracted from similar lip frames into a feature extractor designed to capture spatiotemporal inconsistencies and detect lip-syncing deepfakes. This study introduces LIPINC-V2, building upon our previous work with three key advancements: 1) In mouth inconsistency learning, we enhance our model by integrating a Vision Temporal Transformer, which effectively captures both spatial and temporal information from videos. Furthermore, we incorporate a multi-head cross-attention \cite{vaswani2017attention} module to seamlessly fuse information from RGB mouth sequences and delta frame sequences. These advancements have led to a significant performance improvement across all three datasets. 2) We introduce a segment-wise localization task that segments input videos and classifies each segment as real or deepfake. This approach ensures a comprehensive analysis by examining every part of the video and is particularly effective in detecting localized manipulations. 3) To assess the proposed method's performance in real-world scenarios, we constructed a challenging AI-synthesized lip-syncing dataset, LipSyncTIMIT, comprising lip-syncing deepfakes generated using five state-of-the-art generative models.


The remainder of the article is organized as follows. Related works are briefly reviewed in Section~\ref{sec:Related Work}. The proposed LIPINC-V2 method for lip-syncing deepfake detection is presented in Section~\ref{sec:method}. The LipSyncTIMIT dataset is presented in Section~\ref{sec:experiment}, which also covers the experimental results and detailed analysis. Evaluation on segment-wise localization is provided in Section~\ref{sec:localization}. Lastly, Section~\ref{sec:conclusion} concludes the paper and discusses future work.

\section{Related Work} \label{sec:Related Work}

\noindent{\bf DeepFake Detection}. Deepfakes can be broadly categorized into two types based on the regions of manipulation: entire-face synthesis and partial manipulation. Entire-face synthesis involves techniques like face-swapping and talking head generation, where the entire face is replaced or altered. In contrast, partial manipulation primarily includes lip-syncing deepfakes, where only the lip movements are modified to match a specific audio track. These lip-syncing deepfakes are particularly challenging to detect because they focus on the mouth region while keeping the rest of the face unchanged, often resulting in highly realistic outputs. A study by Sundstrom \etal \cite{sundstrom2023deepfake} highlighted that lip-syncing deepfakes pose a greater challenge for human detection compared to face-swapping deepfakes. The subtlety of the manipulation, confined to the mouth area, makes it difficult for individuals to spot inconsistencies, particularly during real-time playback. Although face-swapping deepfakes have been extensively researched \cite{FTCN,li2018ictu,li2018exposing,nguyen2022deep,muppalla2023integrating}, lip-syncing deepfakes remain a less explored domain. Detecting these manipulations requires identifying subtle audio-visual mismatches and spatial-temporal inconsistencies, which may not be evident in individual frames.

Most deepfake detection methods target full-face synthesis in videos, focusing on identifying visual anomalies such as face warping~\cite{li2018exposing}, irregular facial movements~\cite{trinh2021interpretable}, discrepancies between the face and its context~\cite{nirkin2021deepfake}, and inconsistencies in facial identity~\cite{huang2023implicit}. These approaches extract features from the whole face but struggle with partially manipulated lip-sync videos. For lip-syncing deepfake detection, several techniques exploit audio-visual mismatches. Shahzad \etal~\cite{LipSyncMatters} proposed detecting fabricated videos by contrasting real lip sequences from video clips with synthetic lip sequences generated from audio using a lip-syncing model~\cite{lipsync}. Feng \etal~\cite{feng_et_al} investigated audio-visual signal inconsistencies in forged videos through anomaly detection.  Oorloff et al. \cite{oorloff2024avff} introduces a two-stage deepfake detection framework that captures audio-visual correspondences to identify deepfakes. The first stage involves self-supervised representation learning using real videos to establish natural audio-visual relationships through contrastive learning and complementary masking strategy. In the second stage, these learned features are fine-tuned using supervised learning for deepfake classification. Haliassos \etal~\cite{Lips-don't-lie} identified forged videos by detecting semantic anomalies in mouth movements, focusing on irregularities in motion within lower face regions. It uses a spatio-temporal network pretrained on lipreading tasks to learn natural mouth movement patterns. A temporal network  is then fine-tuned to detect deepfakes by identifying irregularities in mouth motion.  Tolosana \etal~\cite{tolosana2021deepfakes} explored visual artifacts across facial regions (eyes, nose, and mouth) and utilized spatial indicators to differentiate real from fake videos.

Recent studies have advanced by analyzing the temporal correlation between lip movements and audio signals. Liu \etal~\cite{liu2024lips} created a lip-forgery detection method by capturing temporal inconsistencies between lip and head motions with respect to audio signals. This approach also mimics human cognition by identifying subtle biological correlations between the lips and head movements to boost performance. Yu \etal~\cite{yu2024explicit} introduced a correlation distillation task that explicitly learns cross-modal correlations to enhance detection across diverse deepfake scenarios. Yang \etal~\cite{yang2023avoid} made further contributions with AVoiD-DF, an audio-visual joint learning framework that integrates audio and visual features in a complementary manner. This multi-modal approach significantly improves the generalizability of the deepfake detector.

\noindent{\bf Lip-syncing Deepfake Datasets}. Early deepfake datasets are primarily focused on entire-face synthesis, such as face-swapping and face-reenactment~\cite{Faceforensics++,Celeb-df}. However, with the advent of Wav2Lip~\cite{lipsync} in 2020, lip-syncing deepfakes gained prominence, leading to the development of several datasets tailored for lip-syncing detection. The FakeAVCeleb \cite{FakeAVCeleb} dataset, for instance, is a comprehensive audio-video deepfake detection resource. It consists of three categories of audio-video deepfakes, including lip-syncing, face-swapping \cite{faceswap,Fsgan}, and voice cloning \cite{voice-clones}. The KODF \cite{Kodf} dataset also presents a diverse collection of real and fake videos of Korean subjects, generated using six different deepfake generation models \cite{lipsync,DeepFaceLab-paper,Fsgan,Faceswap2020,siarohin2019first,yi2020audio}.
Another notable contribution comes from the CMDFD dataset~\cite{yu2024explicit}, which includes diverse cross-modal forgeries like lip-syncing and talking-head generation.
Liu \etal \cite{liu2024lips}  also curated a high-quality lip-syncing dataset with state-of-the-art generators, further advancing research in this challenging domain. Their dataset includes nearly 340,000 samples.

\begin{figure*}[t]
\centering
\includegraphics[width=\textwidth]{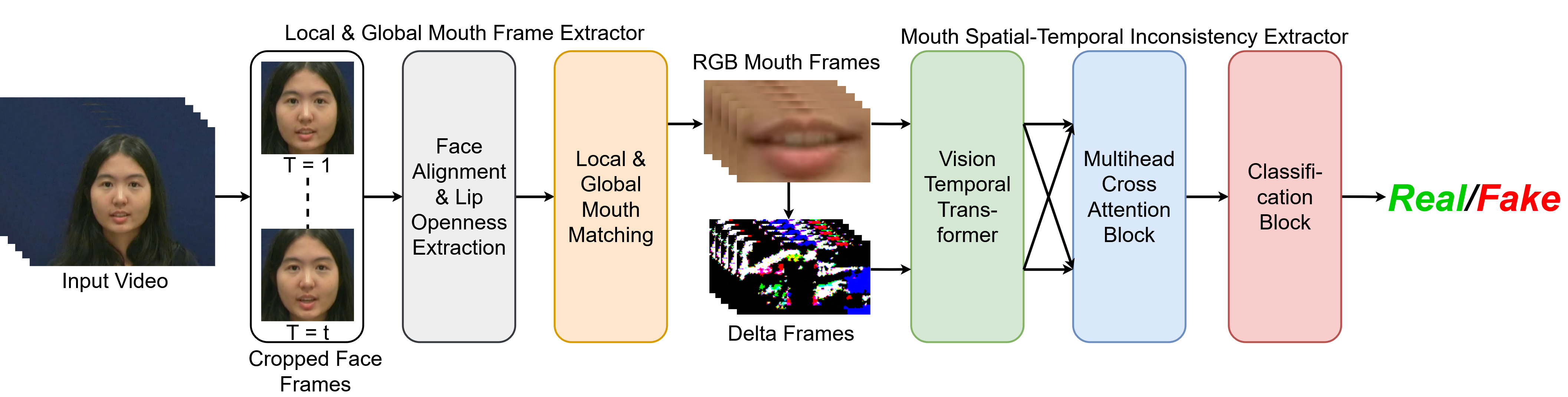}
\caption{\em \small End to End pipeline of the proposed LIPINC-V2 model. Our approach comprises two main modules: (1) Local and Global Mouth Frame Extractor, responsible for isolating adjacent, and similarly posed mouth segments based on mouth openness over time; and (2) Mouth Spatial-Temporal Inconsistency Extractor, tasked with learning distinctive inconsistency features both within and across frames by leveraging mouth appearance and delta frames. }
\label{fig:LIPINC}
\end{figure*}

As deepfake creation techniques evolve, the need for comprehensive datasets that cover a wide range of forgery types becomes more critical. Existing datasets like FaceForensics++ \cite{Faceforensics++} and Celeb-DF \cite{Celeb-df} focus on facial manipulations, but lip-syncing datasets like FakeAVCeleb \cite{FakeAVCeleb}, and KODF \cite{Kodf} are crucial for advancing lip-syncing detection. Moreover, Yang \etal \cite{yang2023avoid}  emphasize the importance of multi-modal datasets that integrate both audio and visual cues to improve detection performance. Their DefakeAVMiT dataset includes multiple perturbation techniques, making it highly relevant for real-world applications where deepfakes may be exposed to various transformations.

As the field of deepfake detection progresses, these datasets and approaches collectively contribute to enhancing the robustness and generalizability of detection methods, particularly in the context of lip-syncing. By leveraging spatio-temporal features and cross-modal correlations, future deepfake detectors can be better equipped to handle the complexities of both in-domain and out-of-domain detection challenges.

\section{Method} \label{sec:method}

The diversity in video attributes such as resolution, head poses, facial expressions, and lighting conditions poses significant challenges for developing an effective detection model. Additionally, recent advancements in deepfake technology have made it increasingly difficult to detect inconsistencies within generated videos. To address these complexities, we introduce our model, which is specifically designed to identify spatial-temporal mouth inconsistencies on both local and global scales for deepfake detection. The model evaluates a video to ascertain its authenticity, categorizing each video with a class label $c \in \{0,1\}$, where 0 denotes Fake, and 1 indicates Real.

The complete workflow of our model is depicted in Fig. \ref{fig:LIPINC}. It comprises two primary modules: 1) \textbf{Local and Global Mouth Frame Extractor}, which isolates adjacent and similarly posed mouth segments based on mouth openness throughout the video sequence; and 2) \textbf{Mouth Spatial-Temporal Inconsistency Extractor} (MSTIE), which is responsible for encoding and learning distinctive inconsistency features within and across frames, focusing on both mouth features and delta frames. These features serve as the input to a binary classifier to determine the probabilities of each class.

Our approach is highly capable of detecting various forms of manipulation that are common in lip-syncing deepfake videos, such as subtle inconsistencies in mouth movements that are not easily visible to the human eye.  By focusing on the spatial-temporal aspects of the video, our model has demonstrated its capability to generalize well across different datasets, outperforming several state-of-the-art models.

\subsection{Local and Global Mouth Frame Extractor}

\begin{figure}[h]
\centering
\includegraphics[width=\columnwidth]{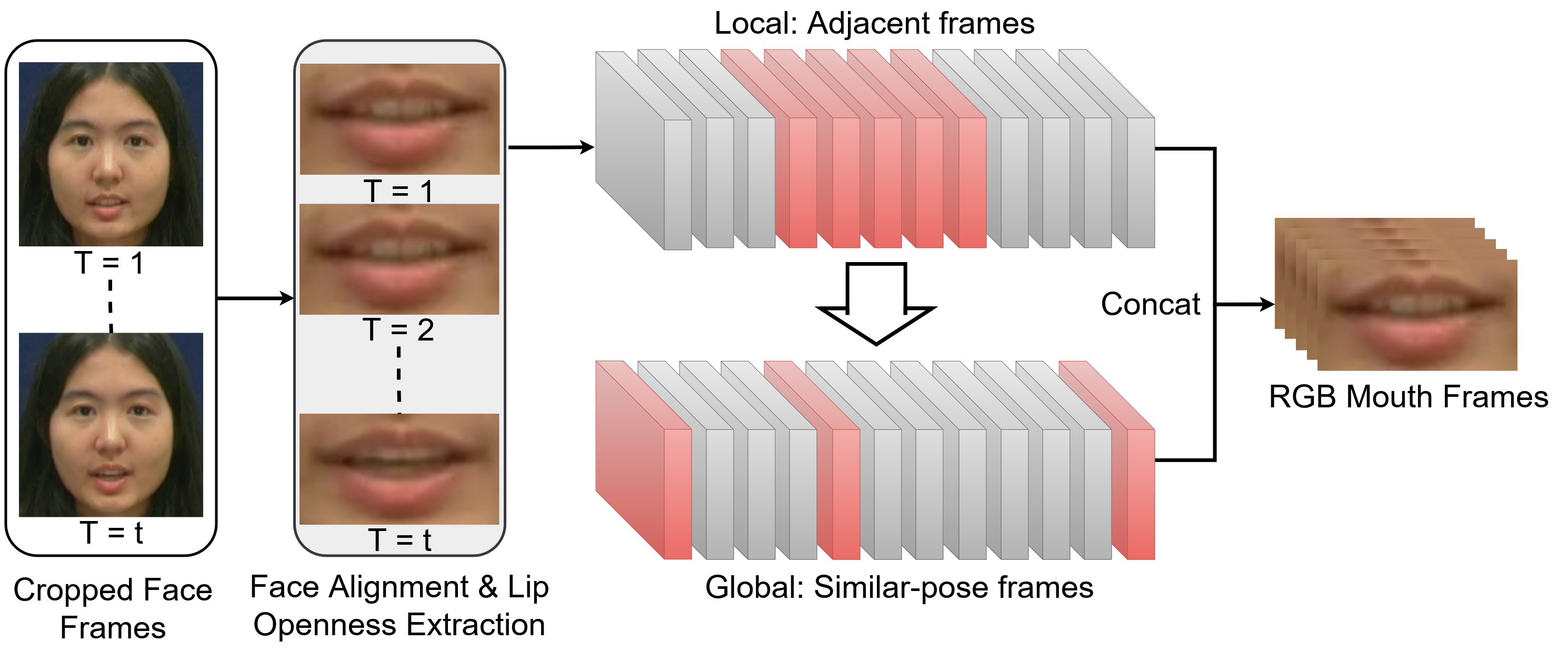}
\caption{Pipeline of the Local \& Global Mouth Frame Extractor. Here T represents frame number. }
\label{fig:LOCALGLOBAL}
\end{figure}

\begin{figure}[h]
\centering
\includegraphics[width=0.5\columnwidth]{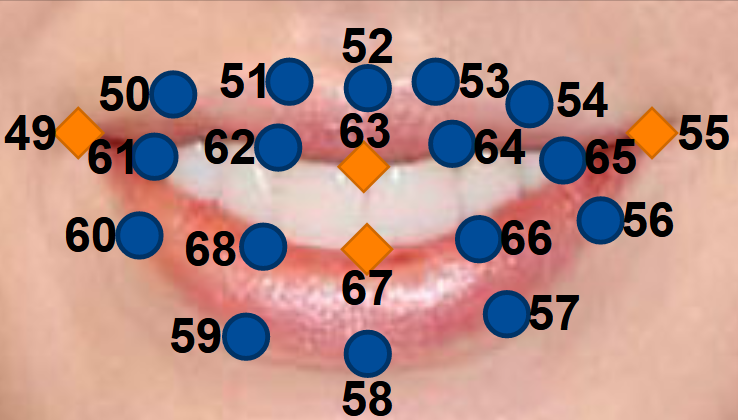}
\caption{ Mouth region landmarks detected by Dlib \cite{Dlib}. Orange colors denote the landmarks for mouth openness measurement and matching.}
\label{fig:dlib}
\end{figure}

The first stage of our method focuses on isolating relevant frames from the video sequence where potential inconsistencies in mouth movement are more likely to occur. As shown in Fig. \ref{fig:LOCALGLOBAL}, given an input video, we first apply a face detector, such as Dlib \cite{Dlib}, to crop and align the face in each frame. This ensures that the facial features, particularly the mouth, are centered in each frame, reducing the influence of irrelevant facial movements or camera angles. Following this, we utilize facial landmarks to precisely extract the mouth region, as illustrated in Fig. \ref{fig:dlib}. 

Our approach emphasizes frames where the mouth is open, as these frames are more likely to contain subtle inconsistencies related to deepfake manipulation. Specifically, we propose the extraction of $L$ adjacent frames and $G$ non-adjacent similar pose mouth frames. The adjacent frames ($L$) capture short-term spatial-temporal relationships, while the non-adjacent similar pose frames ($G$) focus on long-term consistency. These two sets of frames are particularly useful for detecting inconsistencies in lip-syncing videos, where mouth movements may appear natural in individual frames but exhibit irregularities when viewed sequentially.

To identify these frames, we rely on visual landmarks, focusing on the distance between specific points around the mouth (e.g., landmarks 63 and 67 in Fig. \ref{fig:dlib}). By comparing the height and width of the opened lips, along with other facial orientation metrics, we select the $L$ local frames. Then we extract the $G$ global frames that match the degree of mouth openness observed in the local frames. This combination of local and global frame extraction allows the model to detect both short-term and long-term inconsistencies in mouth movement, which are key indicators of deepfake manipulation.

To ensure global matching and avoid selecting adjacent frames, we impose a minimum time gap of 0.09 seconds between the extraction of similar-pose frames. This time gap is based on the average duration of typical speech patterns, ensuring that our model can detect inconsistencies in the mouth over a reasonable time window. We then concatenate the adjacent and similar-pose frames to create our RGB mouth sequence $\textbf{R}$ with $N$ frames where $\textbf{R} \in R^{(N*H*W*3)}$, 3 is the channel number, and $N = L + G$. This approach reinforces our hypothesis that, in a real video, a person's mouth remain consistent throughout.

The  delta frames branch complements this approach by capturing high-level inconsistencies in mouth and teeth shapes across frames. To do this, we compute the residual sequence $\textbf{D}$, where $\textbf{D} \in R^{((N-1)*H*W*3)}$ is calculated by:
\begin{equation}
      D_t = R_{t+1} - R_t \quad (1 \leq t < N)
\end{equation}
The delta frames sequence highlights changes between the consecutive extracted RGB mouth frames, particularly focusing on the edges and contours of the mouth. This structural information is crucial for detecting unnatural mouth movements, which are often overlooked by models that only consider RGB mouth information. Both the RGB mouth and  delta frames sequences are fed into the Mouth Spatial-Temporal Inconsistency Extractor module for feature extraction.

\subsection{Mouth Spatial-Temporal Inconsistency Extractor}

\begin{figure*}[h]
\centering

\includegraphics[width=\textwidth]{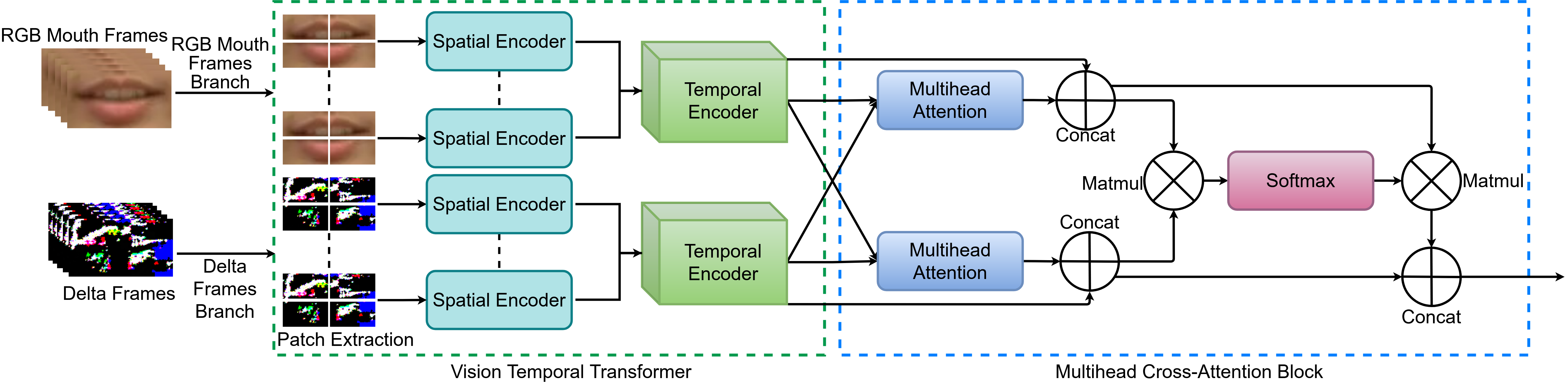}
\caption{The architecture of the Vision Temporal Transformer and the Multihead Cross-Attention Block within our Mouth Spatial-Temporal Inconsistency Extractor. This module encodes RGB mouth and delta frames to learn spatial and temporal inconsistencies using a Vision Temporal Transformer. Through stacked spatial and temporal encoders, it captures relationships within frames and dependencies across frames, focusing on discrepancies in the mouth region. To fully leverage both feature streams, a multi-head cross-attention block interconnects RGB mouth and delta frame branches, ensuring a robust deepfake detection architecture.}
\label{fig:MSTIE}
\end{figure*}

As shown in Fig. \ref{fig:MSTIE} the Mouth Spatial-Temporal Inconsistency Extractor (MSTIE) module is the core of our method. It is designed to encode both the RGB mouth and  delta frames and learn discriminative features that represent spatial and temporal inconsistencies. The inconsistencies we are targeting are not just limited to visual anomalies but also temporal mismatches in how the mouth moves across frames. To achieve this, we utilize a Vision Temporal Transformer(VTT) inspired from modules such as Video Transformer Network \cite{neimark2021video} and Video Swim Transformer \cite{VideoSwimT}, which captures both spatial and temporal information from the video.  It processes input frames by first capturing spatial relationships within individual frames using a spatial transformer encoder and then capturing temporal dependencies across frames using a temporal transformer
encoder. The VTT divides each frame from RGB mouth and  delta frames into non-overlapping patches of size $P \times P$. Each patch is flattened into a vector of dimension $P^2 \cdot 3$, resulting in $X$ patches per frame where $X = \frac{H}{P} \cdot \frac{W}{P}$. Here \(H\) denotes height , \(W\) denotes width , and 3 is the number of channels. A linear projection is then applied to map each patch into a $e$-dimensional embedding space. Positional embeddings are added to these patch embeddings to encode spatial information.

The spatial transformer encoder models relationships between patches within a frame by utilizing multi-head self-attention and feed-forward network. This process is repeated for each spatial transformer layer, capturing intra-frame spatial dependencies. The outputs of the spatial transformer encoder are reshaped into a sequence of frame embeddings by aggregating patch-level features using global average pooling. 
The frame embeddings are then fed into the temporal transformer encoder. This encoder captures inter-frame temporal dependencies. Each layer of the temporal encoder functions similarly to the spatial encoder but processes the sequence of frame embeddings instead.

The Vision Temporal Transformer extracts spatial-temporal features $\textbf{v}_{Ri}$ and $\textbf{v}_{Di}$ for the RGB mouth and  delta frame branches, respectively. Here $i$ refers to the $i$th input video. However, to fully exploit the relationship between RGB mouth and  delta frames, we apply a multi-head cross-attention \cite{vaswani2017attention} mechanism with 4 heads to interconnect the RGB mouth and  delta frame branches. Multi-head attention enables the model to focus on different aspects of the data simultaneously, ensuring that we capture all relevant inconsistencies.

For the RGB mouth branch, we set $K_{Ri} = V_{Ri} = \textbf{v}_{Ri}$ and $Q_{Ri} = \textbf{v}_{Di}$, while for the  delta frames branch, we set $K_{Di} = V_{Di} = \textbf{v}_{Di}$ and $Q_{Di} = \textbf{v}_{Ri}$. The attention scores are then computed for both branches as follows:

\begin{equation}
A_{Ri} = \text{softmax}\left(\frac{Q_{Ri}K_{Ri}^T}{\sqrt{d_{qR}}}\right)V_{Ri}
\end{equation}

\begin{equation}
A_{Di} = \text{softmax}\left(\frac{Q_{Di}K_{Di}^T}{\sqrt{d_{qD}}}\right)V_{Di}
\end{equation}

Next,  a concatenation layer is added after each multihead attention module, the output from the RGB mouth branch is $A_{Ri}+V_{Ri}$ and the  delta frames branch is $A_{Di}+V_{Di}$. These two branches are then fused together using another multi-head cross-attention module. This allows the model to properly learn both the RGB mouth and  delta frames features, focusing more on the  delta frames branch since it results in better generalization performance. The fused attention output is calculated as:

\begin{equation}
A_{fi} = \text{softmax}\left(\frac{Q_{i}K_{i}^T}{\sqrt{d_Q}}\right)V_{i} + Q_{i}
\end{equation}
where $K_{i} = V_{i} = A_{Ri}+V_{Ri}$ and $Q_{i} = A_{Di}+V_{Di}$. The fused feature vector is then passed through a binary classifier to predict whether the input video is real or fake.

\subsection{Loss Function}

To optimize the feature learning process, we design an inconsistency loss $L_{IL}$ to measure the inconsistency level between frames.   We compute the feature vector $F$ from the first layer of our MSTIE module, where $F \in R^{(N'*H'*W'*C')}$ and $N', H', W', C'$ represent the  number of frames, height, width, and channel size, respectively.  For each frame in $F$, we compare it against all other frames in $F$ and calculate the pairwise similarity. The pairwise similarity between frames is calculated using the Structural Similarity Index (SSIM) \cite{SSIM}, and it ranges from 0 to 1, where 1 indicates the frames are similar and 0 indicates the frames are not similar. We then calculate the average similarity score ($AvgS$). For a real video we expect the average similarity score to be close to 1 while for fake videos, it should be near 0. We use a cross-entropy loss function to supervise the average similarity score and compute the inconsistency loss:

\begin{equation}
L_{IL} = - \frac{1}{M} \sum_{i=1}^{M} y_i \log(AvgS_i) + (1 - y_i) \log(1 - AvgS_i)
\end{equation}

Here $M$ is the number of input videos, $y_i \in \{0, 1\}$ where
$y_i = 0$ denotes a fake video and $y_i = 1$ denotes a real video.
In addition to the inconsistency loss, we also apply a classification loss $L_{CL}$ to optimize classification performance. Here we use the binary cross-entropy loss as our classification loss. The total loss function is defined as $L_{total} = \lambda_1 L_{CL} + \lambda_2 L_{IL}$, where $\lambda_1 = 1$ and $\lambda_2 = 5$ are the respective weights for classification and inconsistency losses. Through ablation studies, we find that these values yield the best overall performance.

\section{Experiments} \label{sec:experiment}


\begin{figure*}[t]
\centering
\includegraphics[width=\textwidth]{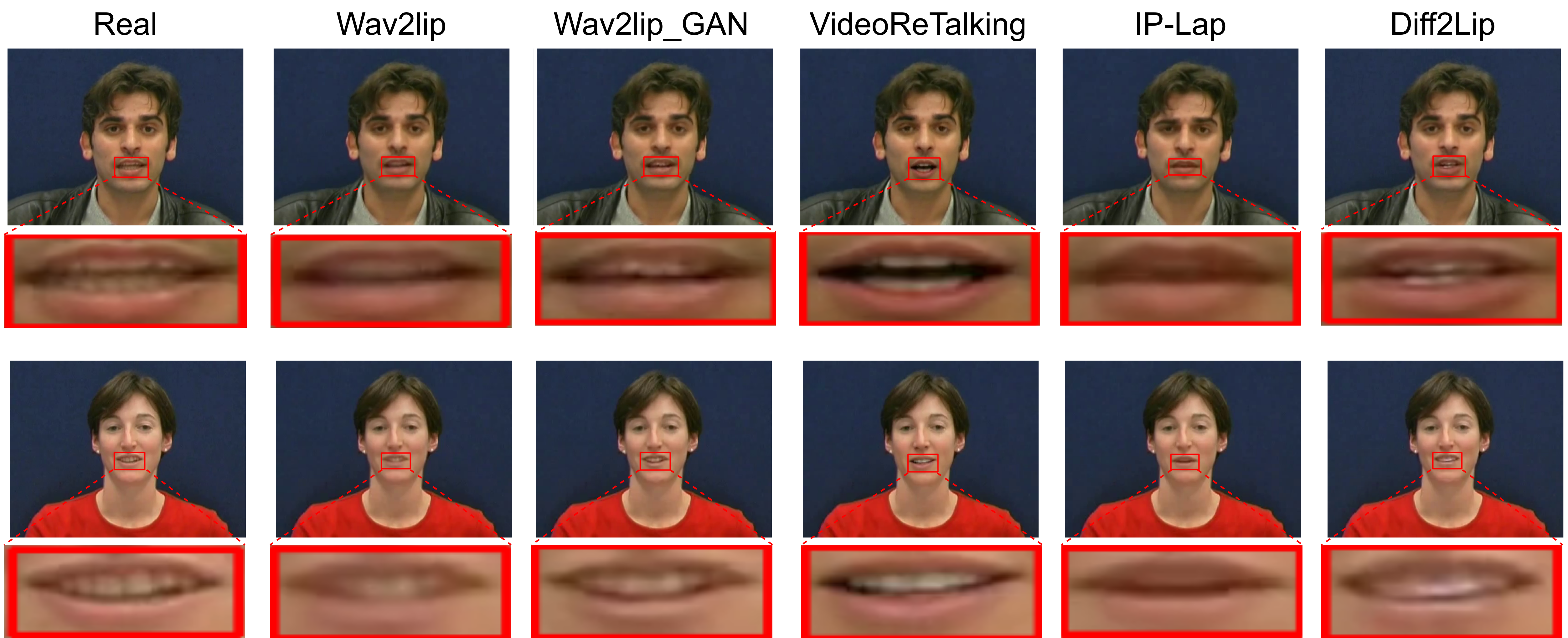}
\caption{Examples of mouth regions from videos generated using five different deepfake generation methods in our LipSyncTIMIT dataset.}
\label{fig:LipSyncTIMIT}
\end{figure*}

\subsection{Datasets}
In this paper, we utilize three datasets to train, test, and validate our model. We introduce our novel dataset, LipSyncTIMIT, alongside two existing multi-modal datasets, specifically designed for lip-syncing deepfake detection, to evaluate the performance of our method.

\vspace{1mm}
\noindent\textbf{LipSyncTIMIT Dataset.}
The exploration of deepfake detection has led to the development of various detection methods and datasets. However, existing deepfake datasets primarily focus on face-swapping and face-reenactment deepfakes, with limited attention given to lip-syncing deepfake generation methods. To address this gap, we introduce the LipSyncTIMIT dataset, designed to enhance research in generalizable detection techniques, particularly in the domain of lip-syncing deepfake generation.

To create this dataset, we sourced real videos from the VidTIMIT \cite{VidTIMIT} dataset, which contains recordings of 43 volunteers reciting short sentences, with synchronized audio and visual tracks. Additionally, real audio samples were collected from the LRS2 \cite{LRS2} dataset, and AI-generated (fake) audio samples were obtained from the LibriSeVoc \cite{sun2023ai} dataset. Inspired by the FakeAVCeleb dataset \cite{FakeAVCeleb}, we created three categories of deepfake videos, as described below:

\begin{itemize}
    \item \textbf{FakeVideo-OriginalAudio:} In this category, the audio from the real videos in the VidTIMIT \cite{VidTIMIT} dataset is used to manipulate the real videos from the same dataset, resulting in lip-syncing deepfakes.
    \item \textbf{FakeVideo-RealAudio:} Here, real audio samples from the LRS2 \cite{LRS2} dataset are used to generate lip-syncing deepfakes by manipulating the real videos from the VidTIMIT \cite{VidTIMIT} dataset.
    \item \textbf{FakeVideo-FakeAudio:} In this case, AI-generated fake audio samples from the LibriSeVoc \cite{sun2023ai} dataset are used to generate lip-syncing deepfakes by manipulating the real videos from the VidTIMIT \cite{VidTIMIT} dataset.
\end{itemize}

We also created two compressed versions of our dataset using constant rate factors of 23 and 40.
 The lip-syncing deepfakes were created using five state-of-the-art lip-syncing deepfake generation methods. In total, our dataset includes 9,090 lip-syncing deepfake videos, along with their compressed versions\footnote{\url{https://github.com/skrantidatta/LIPINC-V2}}. We used five lip-syncing deepfake generation methods to create these deepfake videos.  Fig. \ref{fig:LipSyncTIMIT}  shows how different the mouth looks for different generation methods. The five deepfake generation methods used to create the dataset are as follows:

\begin{itemize}
    \item Wav2lip \cite{lipsync} approach employs a pre-trained lip-synchronization discriminator to direct a generator in generating lip-syncing deepfake videos. The method ensures high-quality output by optimising reconstruction loss and synchronization loss.
    \item Wav2lip\_GAN \cite{lipsync} improves video quality by adding a visual quality discriminator to the Wav2Lip model, but it has a slight trade-off in lip synchronization accuracy.
    \item VideoReTalking \cite{videoretalking} utilizes a three-stage process: expression neutralization, lip-synchronization generation with a pre-processed frame, and identity-aware enhancement, to produce high-quality, synchronized lip-syncing deepfakes.
    \item IP-Lap \cite{IP-LAP} uses a two-stage framework: first, a transformer-based generator predicts facial landmarks from audio, and then a rendering network that converts these landmarks into realistic, identity-preserving lip-syncing deepfake videos.
    \item Diff2Lip \cite{diff2lip} approach inpaints the mouth region with an audio-conditioned diffusion model, resulting in high-quality lip synchronization while retaining identity, posture, and visual consistency across frames.
\end{itemize}

\vspace{1mm}
\noindent\textbf{FakeAVCeleb Dataset.} We utilize the  FakeAVCeleb dataset \cite{FakeAVCeleb} to train and validate our model. This dataset comprises 500 unique frontal face videos of real subjects and over 20,000 multi-modal deepfake videos. The FakeAVCeleb dataset is derived from a curated selection of real YouTube videos from the VoxCeleb2 dataset. They used the Faceswap\cite{faceswap} and the FSGAN\cite{Fsgan} models to create face-swapping deepfakes, the Wav2Lip\cite{lipsync} model to create lip-syncing deepfakes and the SV2TTS\cite{voice-clones} model for voice cloning. For our study, we categorize the dataset into two distinct groups: Lip-syncing deepfakes (FakeAV-LS), created using Wav2Lip \cite{lipsync}, and face-swapping deepfakes (FakeAV-FS), which do not use Wav2Lip\cite{lipsync}. Given that our method primarily targets the detection of lip-syncing deepfakes, we further divided the lip-synced portion of the dataset(FakeAV-LS) into a non-overlapping training set and test set to train and evaluate our model.

\vspace{1mm}
\noindent\textbf{KODF Dataset.}
 The KODF dataset~\cite{Kodf}, includes 62,166 real videos and 175,776 deepfake videos. Similar to the FakeAVCeleb dataset\cite{FakeAVCeleb}, this dataset consists of lip-syncing and face-swapping deepfake videos. We created 3 test sets from the dataset based on the deepfake generation method. 1) KODF-LS: Audio-driven lip-synced deepfakes generated using \cite{lipsync, ATFHP}; 2) KODF-FSGAN: Faceswap deepfakes generated using \cite{Fsgan}; 3) KODF-DFL: Faceswap deepfakes generated using \cite{DeepFaceLab-paper}. We randomly selected a subset of real and fake videos for evaluation following previous studies~\cite{feng_et_al}. 

\noindent\textbf{Evaluation Metrics.} The model is evaluated using three widely used metrics, including  Average Precision (AP) and Area Under the Receiver Operating Characteristic Curve (AUC) scores \cite{AUC} for the binary detection task and Intersection over Union (IoU) for the localization task. 

\subsection{Experimental Settings} In the proposed architecture, we use the Dlib toolkit \cite{Dlib} to extract $64\times144$ mouth regions from the input video. We set the local frame number $L$ as 5 and the global frame number $G$ as 3 such that the model can detect the required number of global frames even if the video is only a few seconds long. For all experiments, we use Adam optimizer\cite{adam} with 0.001 learning rate and 0.1 epsilon. We train our model for 100 epochs with a batch size 64 on the Keras framework with TensorFlow version 2.15.0.

\begin{table}[h]
\centering
\caption{\textbf{Comparison of AP and AUC score across FakeAV-LS test set (in-domain testing) and KODF-LS datasets (cross-domain testing).} \# and + indicate that the results were sourced from the papers \cite{oorloff2024avff} and \cite{feng_et_al}, respectively.
}
\resizebox{\columnwidth}{!}{
\begin{tabular}{l|rr|rr}
    \toprule
    \multirow{2}{*}{Method} & \multicolumn{2}{c|}{FakeAV-LS} & \multicolumn{2}{c}{KODF-LS} \\
    \cmidrule{2-5}
    & AP & AUC & AP & AUC \\
    \midrule
    \textbf{Intra-modal}\cite{tian2023unsupervised}  & 0.94   & 0.67    & -    & -    \\
    \textbf{Intra-Cross-modal}\cite{tian2023unsupervised}  & \textbf{0.99} & 0.96 & -    & -    \\
    \textbf{RealForensics}\cite{RealForensics}\#  & 0.91   & 0.95    & {0.96} & {0.94} \\
    \textbf{LipForensics}\cite{Lips-don't-lie}\#    & 0.98   & 0.98    & 0.90 & 0.87 \\
    \textbf{Xception}\cite{Faceforensics++}\#   & 0.90   & 0.90    & 0.77 & 0.78 \\
    \textbf{FTCN}\cite{FTCN}\#   & 0.96   & 0.97    & 0.67 & 0.68 \\
    \textbf{AV-DFD}\cite{zhou2021joint}\#   & \textbf{0.99}   & \textbf{0.99}    & 0.80 & 0.82 \\
    \textbf{AVAD }\cite{feng_et_al}  & 0.94   & 0.94    & 0.88 & 0.87 \\
    \textbf{VQ-GAN}\cite{esser2021taming}+  & 0.56   & 0.53    & 0.47 & 0.46 \\
    \textbf{AVFF}\cite{oorloff2024avff}  & 0.97   & \textbf{0.99} & 0.93 & 0.96 \\
   \midrule
    \textbf{LIPINC}\cite{LIPINC}(Our)  & 0.94   & 0.97    & 0.90 & 0.91 \\
    \textbf{LIPINC-V2 (Proposed)} & \textbf{0.99} & \textbf{0.99} & \textbf{0.97} & \textbf{0.97} \\
    \bottomrule
\end{tabular}
}

\label{tab:comparison}
\end{table}

\subsection{Comparison under In-domain Testing}

In Table \ref{tab:comparison}, we evaluate our LIPINC-V2 model on the FakeAV-LS test set \cite{FakeAVCeleb} and compare its performance with various state-of-the-art (SOTA) methods. Our proposed model, demonstrates exceptional performance, achieving the highest AP score of 0.99, tied with Intra-Cross-modal \cite{tian2023unsupervised} and AV-DFD \cite{zhou2021joint}. This highlights the model's capability to accurately classify lip-syncing deepfake videos. Furthermore, our model achieves a significant result in terms of AUC, scoring 0.99, which is on par with the AVFF \cite{oorloff2024avff} and AV-DFD \cite{zhou2021joint}.

Compared to our previous model LIPINC \cite{LIPINC}, which achieved 0.94 in  AP and 0.97 in AUC, our proposed model shows a significant improvement. We have improved the performance by 5\% in terms of AP and 2\% in terms of AUC.  Our LIPINC-V2 model's superior performance as compared to state-of-the-art methods  can be attributed to the effective integration of Vision Temporal Transformer with Multihead Cross-Attention, which enables the model to better capture spatial-temporal features. This is crucial for identifying subtle inconsistencies in lip movements, thereby improving the detection of lip-syncing deepfake videos.


\subsection{Comparison under Cross-domain Testing}
We further evaluate our model on the KODF-LS \cite{Kodf} and LipSyncTIMIT datasets to assess its generalization capabilities.

\noindent\textbf{Analysis on  KODF-LS \cite{Kodf} dataset.} 
The results are presented in Table \ref{tab:comparison}. It is important to note that our  model was trained exclusively on the FakeAV-LS training set , while the state-of-the-art (SOTA) methods, referenced from existing papers, were trained on the entire FakeAVCeleb dataset \cite{FakeAVCeleb} as detailed in \cite{feng_et_al}.
On the KODF-LS dataset, our model again shows strong performance, achieving an AP of 0.97 and an AUC of 0.97, outperforming the best model in terms of AP,  Real Forencics \cite{RealForensics} model  by 1\% in terms of AP and 3\% in terms of AUC. We outperform the best model in terms of AUC, AVFF \cite{oorloff2024avff} model by 1\% in terms of AUC and 4\% in terms of AP. Despite not being trained on KODF-LS data, our model performs well, demonstrating its strong generalization ability to different datasets.

\begin{figure*}[t]
\centering
\includegraphics[width=\textwidth]{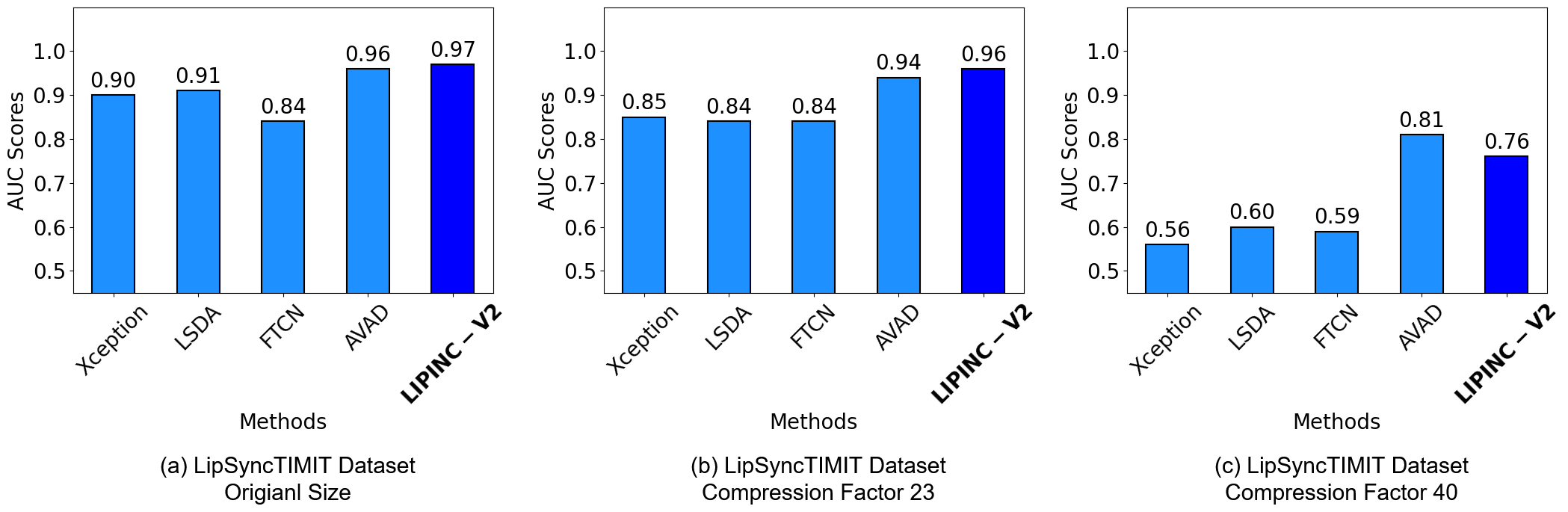}
\caption{ Comparison with state of the art models under cross-domain testing on different compression versions of the LipSyncTIMIT dataset.}
\label{fig:robust2}
\end{figure*}

\noindent\textbf{Analysis on  LipSyncTIMIT dataset.} 
Fig. \ref{fig:robust2} illustrates the evaluation of several detection models on the proposed LipSyncTIMIT dataset, including its compressed versions with  constant rate factor of 23 and 40. It can be observed that, our proposed model achieved the best performance with an AUC of 0.97 on uncompressed data, outperforming all the baselines. This demonstrates that our model effectively generalizes to new datasets containing diverse types of lip-syncing manipulations. When tested on the compressed dataset with a constant rate factor of 23, all models exhibited minimal performance degradation, with the LSDA \cite{lsda} model experiencing the highest decline at 7\%. When the dataset was compressed to a factor of 40, the performance of all models dropped significantly to approximately 0.60 AUC, except for our LIPINC-V2 model and AVAD \cite{feng_et_al}. Unlike most SOTA models, our model demonstrated superior robustness, achieving an AUC score of 0.76, only outperformed by AVAD\cite{feng_et_al} with 0.81 AUC. The ability of our model to handle unseen lip-syncing deepfakes, even under heavy compression, underscores the effectiveness and robustness of the spatial-temporal features it captures.

In summary, our model achieves high AUC scores on unseen datasets, such as LipSyncTIMIT and KODF-LS, with AUC scores of 0.97 on both datasets. These results validate its capability to generalize beyond the training data, highlighting its potential for detecting lip-syncing deepfakes in cross-domain settings.

\begin{figure}[t]
\centering
\includegraphics[height=0.65\columnwidth]{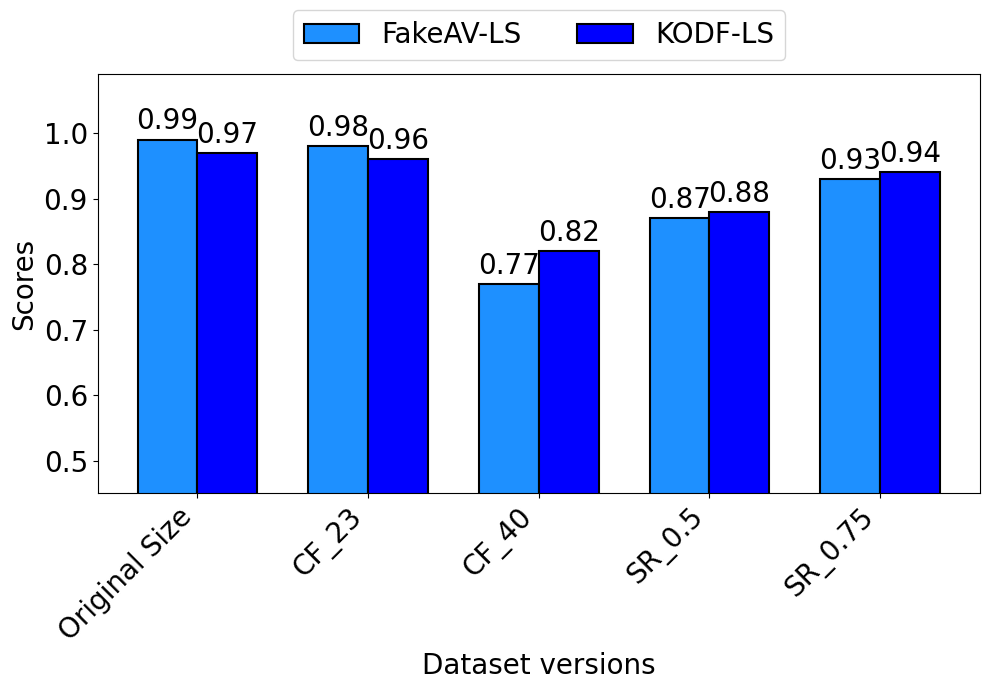}
\caption{AUC scores for different compression and size reduction methods on FakeAV-LS test set and KODF-LS datasets. Here CF\_23 and CF\_40 represent videos compressed with constant rate factors of 23 and 40, respectively, while SR\_0.5 and SR\_0.75 indicate videos scaled down to 50\% and 75\% of their original resolution, respectively.
}
\label{fig:robust}
\end{figure}

\subsection{Robust Testing}
To evaluate the robustness of our model, we tested its performance on various compressed versions and size reductions using the FakeAV-LS test set and KODF-LS dataset. The model was trained on the FakeAV-LS training set and evaluated on four versions of the FakeAV-LS test set and KODF-LS dataset. These versions include two compression versions denoted as CF\_23 and CF\_40, and two resolution reductions SR\_0.5 and SR\_0.75. For CF\_23 and CF\_40, the videos were compressed with constant rate factors of 23 and 40, respectively. For the resolution reductions, the videos were scaled down to 50\% and 75\% of their original size .

As shown in Fig.\ref{fig:robust}, the model demonstrated strong resilience to various compression and resolution modifications. Our model is able to perform well in almost all cases, achieving an AUC score of 0.82 or higher. The most significant performance drop occurred with CF\_40 compression, where the AUC decreased by 22\% for FakeAV-LS test set and by 15\% for KODF-LS dataset. Similarly, with SR\_0.5 resolution reduction, the AUC decreased by 12\% for FakeAV-LS test set, while AUC decreased by 9\% for KODF-LS. For SR\_0.75, the AUC decreased by 6\% for FakeAV-LS test set, while KODF-LS remained relatively stable with only a 3\% decline in AUC. In contrast, the model exhibited minimal performance loss with CF\_23, showing only a 1\% reduction in AUC on both FakeAV-LS test set and KODF-LS dataset.


These findings emphasize the model's practical reliability and efficiency even under significant video compression and quality variations. Its ability to maintain good AUC scores demonstrates the robustness of its prediction capabilities in real-world scenarios where media quality may fluctuate due to compression or downsampling. The model's resilience ensures it remains a reliable tool for lip-syncing deepfake detection.

\subsection{Ablation Studies}
In this subsection, we present two sets of ablation analysis. The first involves integrating audio with the frames of the video as input, while the second focuses on the impact of each component in our model. The first set experimental results are visualized in Table~\ref{tab:comparisonAudio}. Here we show the detection capabilities of our model on three datasets, namely FakeAV-LS test set \cite{FakeAVCeleb}, KODF-LS \cite{Kodf} and the proposed LipSyncTIMIT dataset. The second set   of  the experimental results are visualized in Fig. \ref{fig:ablation}. The performance of different model configurations, including variants without key components, is analyzed to investigate the detection capabilities of the proposed architecture on our  proposed LipSyncTIMIT dataset.

\subsubsection{Integrating Audio as Input}
As shown in Table \ref{tab:comparisonAudio}, we experimented with adding audio input in two different parts of the model:

\begin{table}[t]
\centering
\caption{\textbf{Ablation Analysis of AP and AUC score across FakeAV-LS test set, LipSyncTIMIT, and KODF-LS datasets. Here Audio\_LGMFE refers to the integration of audio into the Local and Global Mouth Frame Extractor module, while Audio\_MSTIE represents the integration of audio into the Mouth Spatial-Temporal Inconsistency Extractor.
}}
\begin{tabular}{l|rr|rr|rr}
    \toprule
    \multirow{2}{*}{Method} & \multicolumn{2}{|c|}{FakeAV-LS} & \multicolumn{2}{c|}{LipSyncTIMIT} & \multicolumn{2}{c}{KODF-LS} \\
    \cmidrule{2-7}
    & AP & AUC & AP & AUC & AP & AUC \\
    \midrule
    \textbf{Audio\_LGMFE}   & 0.98 & 0.94 & 0.53 & 0.68 & -    & -    \\
    \textbf{Audio\_MSTIE}      & 0.96 & 0.85 & 0.52 & 0.63 & 0.68 & 0.68 \\
    \textbf{LIPINC-V2}         & \textbf{0.99} & \textbf{0.99} & \textbf{0.98} & \textbf{0.97} & \textbf{0.97} & \textbf{0.97} \\
    \bottomrule
\end{tabular}

\label{tab:comparisonAudio}
\end{table}

\noindent\textbf{Integrating Audio to the Local and Global Mouth Frame Extractor Module (Audio\_LGMFE).} In this approach, we integrate both phonemes and visual landmarks to enhance lip-syncing deepfake detection accuracy by leveraging a dual-modality framework. Using the Whisper~\cite{bain2022whisperx} model, we extract the video transcript along with precise timestamps. From the transcript, phonemes are generated using the g2p \cite{g2pE2019} model, prioritizing those associated with open-mouth movements (e.g., \textit{o}, \textit{a}, \textit{e}). To align phonemes with visual data, frames depicting similar poses and phonemes are identified by analyzing the dimensions and proportions of the opened lips relative to comparable facial orientations. We use this process  to extract $G$ similar-pose frames. This combination of audio and visual features refines the frame selection process, ensuring that chosen frames accurately represent the intended synchronization. A single frame is selected from the refined similar-pose frame set $G$, and $L$ adjacent frames to the selected frame are extracted. As shown in Table~\ref{tab:comparisonAudio}, this method, denoted as \textit{Audio\_LGMFE}, achieves an AUC of 0.94 on the FakeAV-LS test dataset, 5\% lower than our proposed approach. On LipSyncTIMIT dataset, it is not able to generalize to a new dataset and both the AP and AUC falls by 45\% and 29\% respectively, compared to our proposed model. This approach faces challenges with non-English datasets such as KODF \cite{Kodf} due to the limitations of the g2p\cite{g2pE2019} model and struggles to outperform our proposed method.

\noindent\textbf{Integrating Audio to the Mouth Spatial-Temporal Inconsistency Extractor (Audio\_MSTIE).}
In this approach, we convert the video's audio into Mel-Frequency Cepstral Coefficients (MFCC) features and extract these using a set of 1D convolution layers. These features are then concatenated with the output of the multihead cross-attention module. As shown in Table \ref{tab:comparisonAudio}, this model, referred to as \textit{Audio\_MSTIE}, performs poorly compared to the other two approaches. The reason is that, in this method, when the audio in the deepfake video is real and the video is fake, audio has no relevance to the detection task, thereby negatively impacting model performance. This model shows a 14\% drop in AUC for FakeAV-LS test set and a 34\% drop in AUC for LipSyncTIMIT dataset and a 29\% drop in AUC for KODF-LS compared to our proposed LIPINC-V2 approach.

\subsubsection{Impact of Each Component in Our Proposed Approach}
\begin{figure}[t]
\centering
\includegraphics[width=\columnwidth]{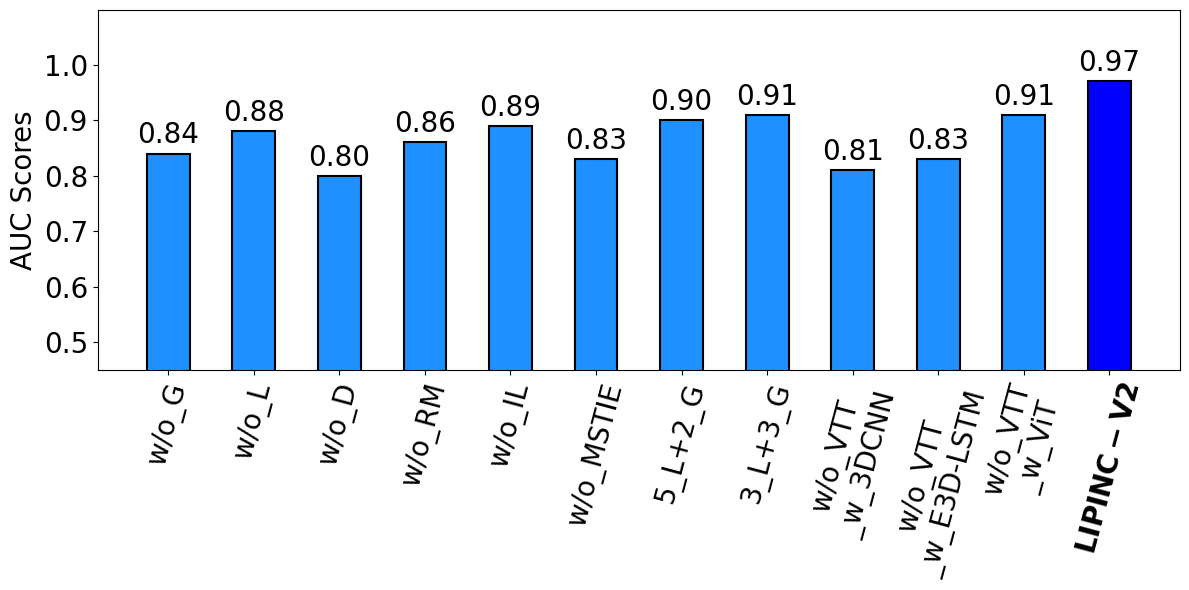}
\caption{ Ablation analysis on  LipSyncTIMIT dataset based on AUC scores. Here  L, G, RM, and D refer to Local, Global, RGB Mouth, and Delta frames, respectively. IL is Inconsistency Loss, MSTIE is  Mouth Spatial-Temporal Inconsistency Extractor and VTT refers to Vision Temporal Transformer. 3DCNN, E3D-LSTM and ViT refers to 3D Convolutional Neural Network\cite{3D_CNN}, Eidetic 3D LSTM \cite{wang2019eidetic} and Vision Transformer\cite{VIT} respectively.
}
\label{fig:ablation}
\end{figure}

We show how each module affects the performance of our model in Fig. \ref{fig:ablation}.

\vspace{5mm}
\noindent\textbf{Effectiveness of Local and Global Frames.} 

Both local and global frames form the foundation of our model. When local frames (\texttt{w/o\_L}) were removed, there was a $9\%$ decrease in AUC on the LipSyncTIMIT dataset. On the other hand, removing global frames (\texttt{w/o\_G}) led to an even more significant decline in performance, with a $13\%$ reduction on the LipSyncTIMIT dataset.

In our proposed approach, we use $5$ local frames and $3$ global frames. We experiment with these hyper-parameters to evaluate how they affect our model's performance. When we removed just one global frame (denoted as $5\_L + 2\_G$), the performance decreased by $7\%$ on the LipSyncTIMIT dataset. For $3\_L + 3\_G$, where we remove two local frames, the performance dropped by $6\%$ on the LipSyncTIMIT dataset.

This underscores the greater impact of global frames on the overall performance of the model.

\noindent\textbf{Effectiveness of RGB Mouth and  Delta Frames.} 
Both RGB Mouth (RM) and  Delta (D) frames contribute significantly to deepfake detection in our model. Removing any of these components also results in the removal of our MSTIE module, since it is used to fuse the RM and D features. Specifically, removing  Delta frames (\texttt{w/o\_D}) led to the largest performance decrease, with a $17\%$ drop in AUC on the LipSyncTIMIT dataset. In contrast, removing RGB Mouth frames (\texttt{w/o\_RM}) resulted in a slightly smaller decrease of $11\%$ on LipSyncTIMIT. These results demonstrate that both RGB Mouth and  Delta frames critically influence the model's overall performance.

\noindent\textbf{Effectiveness of MSTIE Module.} 
The MSTIE module, which fuses RGB Mouth and Delta features, provides significant performance improvements. Removing this module by itself, as seen in the \texttt{w/o\_MSTIE} configuration, resulted in a $14\%$ drop in AUC on the LipSyncTIMIT dataset. This demonstrates the importance of the MSTIE module for achieving high detection accuracy.

\noindent\textbf{Effectiveness of Inconsistency Loss (IL).} 
The inconsistency loss (IL) is crucial for capturing subtle temporal inconsistencies. As demonstrated in the ablation studies, removing the IL (\texttt{w/o\_IL}) caused an $8\%$ drop in AUC on the LipSyncTIMIT dataset. This highlights the importance of the IL in distinguishing deepfakes from real videos and improving the generalization capability of the model.

\begin{figure*}[t]
\centering
\includegraphics[width=0.85\textwidth]{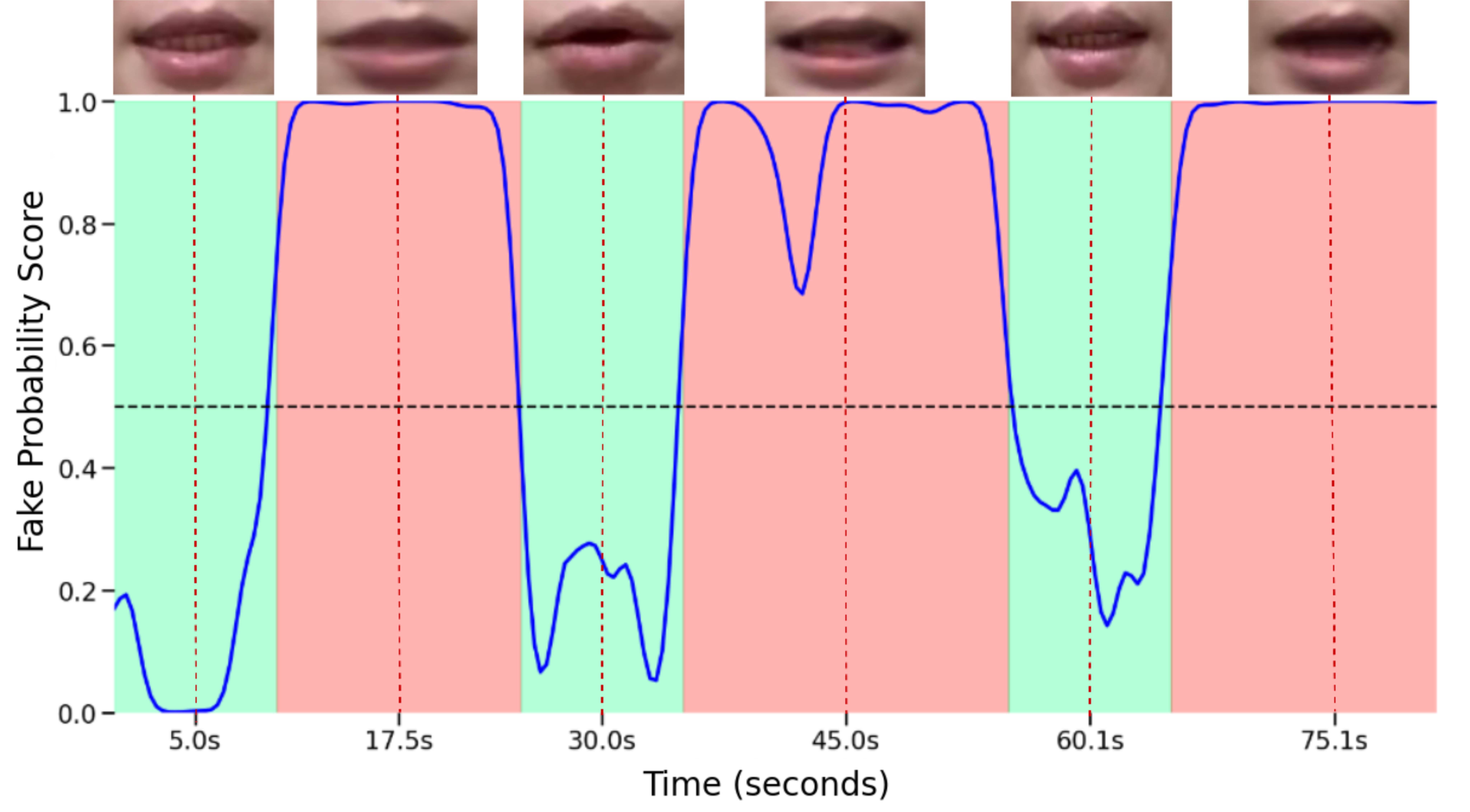}
\centering
\caption{ Example of Segment-wise classification results.
This plot illustrates the predicted fake probability score over time for a sample video. Real segments are denoted in green, while fake (deepfake) segments appear in red. The black dashed line at y=0.5 indicates the decision threshold used to distinguish real from fake. The vertical red dashed line marks the frame of interest at the corresponding position on the time axis.}
\label{fig:Segment}
\end{figure*}

\noindent\textbf{Using 3DCNN in place of Vision Temporal Transformer.} 
In this study, we employed a 3D Convolutional Neural Network (3D-CNN)\cite{3D_CNN} as an alternative to the Vision Temporal Transformer (VTT) in the MSTIE module, denoted as \texttt{w/o\_VTT\_w\_3DCNN} in Fig.~\ref{fig:ablation}. The 3D-CNN utilizes 3D convolutions to extract spatiotemporal features from the input data. 


On the LipSyncTIMIT dataset, the performance of the \texttt{w/o\_VTT\_w\_3DCNN} method showed a $16\%$ decrease in AUC compared to the LIPINC-V2 approach.

\noindent\textbf{Using E3D-LSTM in place of Vision Temporal Transformer.}
Here we used the E3D-LSTM architecture\cite{wang2019eidetic} instead of the Vision Temporal Transformer, denoted as \texttt{w/o\_VTT\_w\_E3D-LSTM} in Fig.~\ref{fig:ablation}. The E3D-LSTM integrates 3D convolutions\cite{3D_CNN} into the LSTM \cite{hochreiter1997lstm} architecture. 
By combining the strengths of 3D convolutions for local motion encoding and LSTMs for  long-term temporal information, the E3D-LSTM architecture provides a comprehensive approach for processing spatiotemporal features in videos.

On the LipSyncTIMIT dataset, we found that the model performs $14\%$ worse in terms of AUC compared to our proposed approach. We believe this is because the E3D-LSTM model  overfits to the training data.

\noindent\textbf{Using Vision Transformer in place of Vision Temporal Transformer.}
In this study, we employed the Vision Transformer (ViT)\cite{VIT} as an alternative to the Vision Temporal Transformer (VTT) in the MSTIE module, denoted as \texttt{w/o\_VTT\_w\_ViT} in Fig.~\ref{fig:ablation}. By removing the temporal dimension, the transformer can no longer capture temporal information, resulting in a reduction in performance. On the LipSyncTIMIT dataset, the performance drops by $6\%$ in terms of AUC compared to our proposed approach.

Overall, the results demonstrate that integrating local frames, global frames, RGB mouth features, and  Delta Frames, combined with the MSTIE module comprising a Vision Temporal Transformer and inconsistency loss, provides the most robust and accurate system for detecting lip-syncing deepfakes.

\subsection{Limitations}
Although our model demonstrates outstanding performance in detecting lip-syncing deepfakes, it has some limitations in practical applications. For example, it cannot detect deepfake videos without visible lip movement. Moreover, the model may struggle with very short videos, where identifying global frames effectively becomes challenging. When tested on face-swapping deepfake datasets, the model, trained solely on FakeAV-LS training set, achieves  0.72 AUC on FakeAV-FS, 0.50 AUC on KODF-FSGAN, and 0.66 AUC on KODF-DFL. These datasets involve face-swapping deepfakes with facial artifacts significantly differing from those in lip-syncing videos, highlighting the need for further adaptation or additional training to handle face-swapping deepfakes effectively.
In future work, we aim to improve the model's generalization across various types of deepfakes by incorporating more global features.

\section{Segment-Wise Localization} \label{sec:localization}

In real-world applications, videos can be partially manipulated, with only specific segments being altered. To address this, we evaluated our model to include a localization task that identifies the specific areas affected by lip-syncing deepfakes. Instead of taking the whole video as a single input, as done in detection tasks, we divide the video into smaller 1-second segments and run the model on each segment to obtain the probability of it being fake. 
This approach ensures that no part of the video is left unanalyzed, making it particularly useful for localizing partial manipulations. We show a localization example in Fig. \ref{fig:Segment}. For each segment, the fake probability score is computed, and the probability of a given frame is directly associated with the fake probability score of the segment to which it belongs. 
This visual representation highlights how the segment-wise model can distinctly separate real and fake segments, even within the same video. The model's sensitivity to subtle manipulations is particularly evident when small, localized alterations in the fake parts are successfully flagged by the model.

In addition to qualitative results, we present quantitative performance metrics in Table \ref{tab:segmentmodel}. The segment-wise model was evaluated on two modified deepfake detection datasets: LipSyncTIMIT-SW and KODF-SW. These datasets were created by concatenating real and fake segments from multiple real and fake videos while ensuring that the face identity remained consistent within a single video. 

Our model performs on par with the best segment-wise baseline models on the KODF-SW dataset and surpasses them on the LipSyncTIMIT-SW dataset. On the KODF-SW dataset, which contains lip-syncing manipulations generated by two models, our model achieved an IoU of 0.89. On the LipSyncTIMIT-SW dataset, which includes lip-syncing manipulations from five different generation models, our model achieved an IoU of 0.75. The decrease in performance is attributed to the inclusion of newer lip-syncing generation models that were not part of the training dataset. 
These results highlight the effectiveness of the proposed model in localizing lip-syncing deepfake content, especially in cases where the manipulations are subtle and partial.


\begin{table}[ht]
\centering
\caption{Performance (IoU) of different models in localizing lip-syncing deepfakes on two Segment-Wise (SW) datasets.}
\begin{tabular}{l|c|c}
    \toprule
    {Method} & \multicolumn{1}{c|}{KODF-SW} & \multicolumn{1}{c}{LipSyncTIMIT-SW} \\
    \midrule
    \textbf{Xception \cite{Faceforensics++}} & \textbf{0.93} & 0.69 \\
    \textbf{Altfreezing \cite{wang2023altfreezing}} & 0.84 & 0.51 \\
    \textbf{FTCN \cite{FTCN}} & 0.78 & 0.48 \\
    \midrule
    \textbf{LIPINC-V2 (Proposed)} & 0.89 & \textbf{0.75} \\
    \bottomrule
\end{tabular}

\label{tab:segmentmodel}
\end{table}


\section{Conclusion} \label{sec:conclusion}
In this paper, we have proposed a novel method for detecting lip-syncing deepfakes by identifying spatial-temporal inconsistencies in the mouth region, aided by vision temporal transformer and  multihead cross-attention. These inconsistencies, which are challenging for humans to detect, appears both within adjacent frames and across the entire video. To capture these inconsistencies, we have designed the Mouth Spatial-Temporal Inconsistency Extractor (MSTIE), guided by an inconsistency loss function. By combining both local and global frames, our method effectively distinguishes real from fake videos. Additionally, we created a new lip-syncing deepfake dataset, LipSyncTIMIT, which was generated using five state-of-the-art lip-syncing models.

Experiments conducted on three datasets demonstrate that our model achieves state-of-the-art performance in both in-domain and cross-domain deepfake detection. In future work, we plan to focus on enhancing the model's ability to detect a wider range of face manipulations. Additionally, we aim to explore more effective multi-modality feature learning to further improve deepfake detection. 



\small{
\bibliographystyle{IEEEbib}
\bibliography{TIFS}
}

\vspace{-10mm}

\begin{IEEEbiography}
[{\includegraphics[width=1in,height=1.25in,clip]{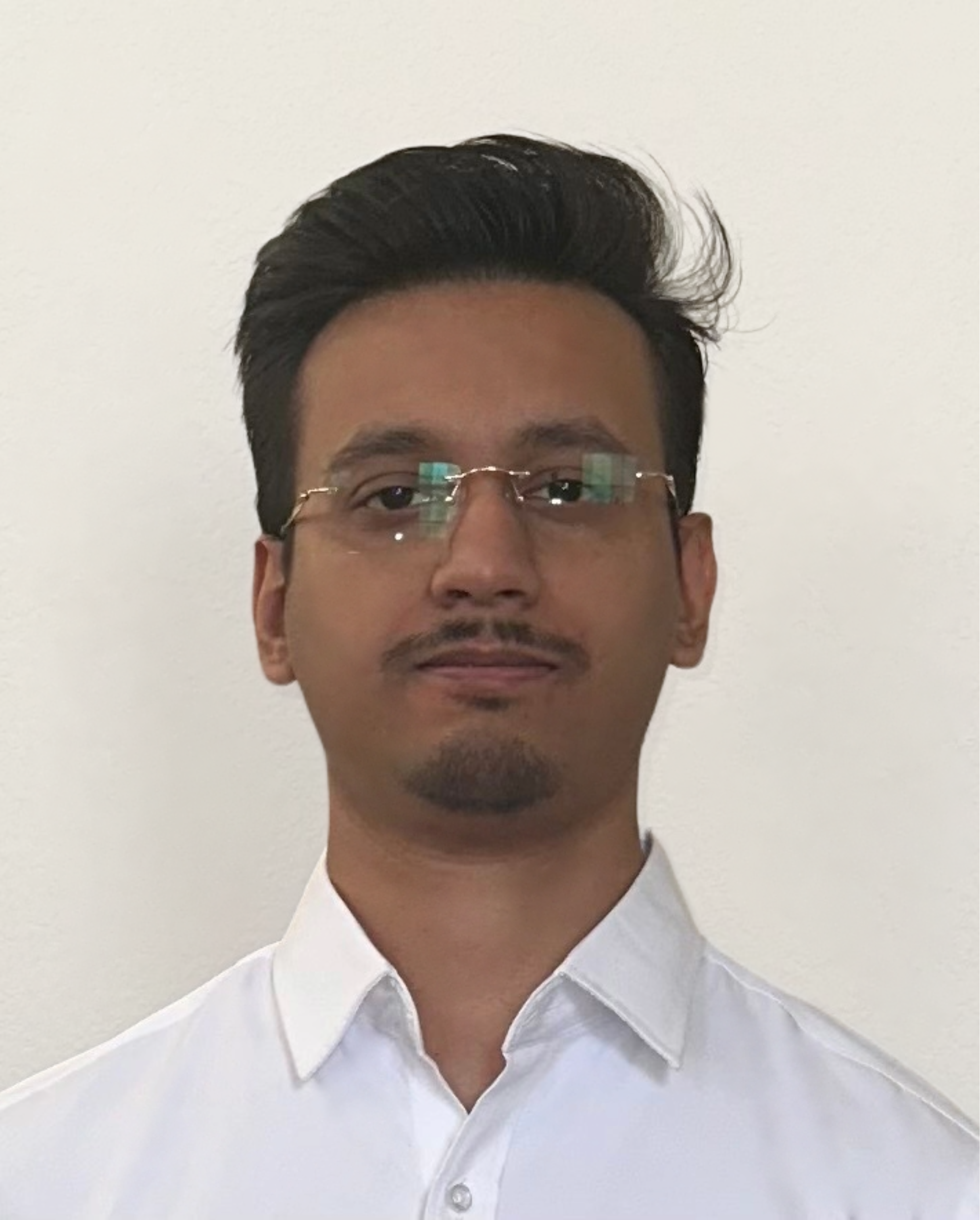}}]{Soumyya Kanti Datta}
received his B.Tech. degree in Computer Science and Engineering from the Kalinga Institute of Industrial Technology, India, in 2018, and his M.S. degree in Computer and Information Sciences from the University at Buffalo, State University of New York, USA, in 2021. He is currently pursuing the Ph.D. degree in the Department of Computer Science and Engineering at the University at Buffalo. His research interests include multimedia forensics and computer vision.
\end{IEEEbiography}
\vspace{-15mm}

\begin{IEEEbiography}
[{\includegraphics[width=1in,height=1.25in,clip]{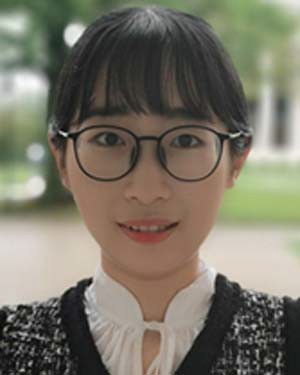}}]{Shan Jia}
received the B.S. degree in electronic and information engineering and the Ph.D. degree in communication and information systems from Wuhan University, Wuhan, China, in 2014 and 2021, respectively. She is a research scientist with the Department of Computer Science and Engineering, State University of New York at Buffalo, Buffalo, NY, USA. Her research interests include multimedia forensics, biometrics, and computer vision.
\end{IEEEbiography}
\vspace{-12mm}

\begin{IEEEbiography}[{\includegraphics[width=1in,height=1.25in,clip]{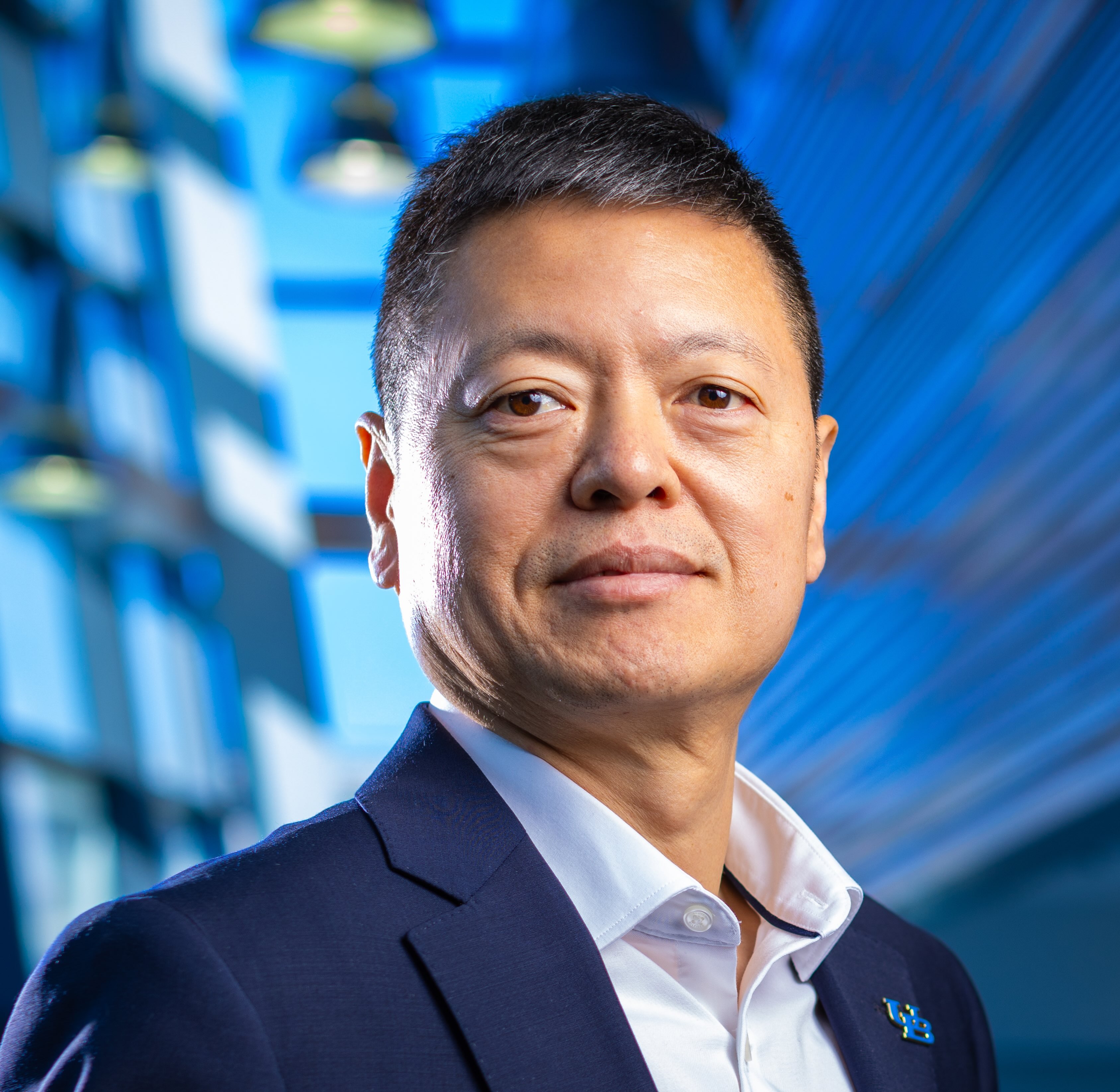}}]{Siwei Lyu} received his B.S. degree in Information Science, and M.S. degree in Computer Science and Technology from Peking University, China, in 1997 and 2000, respectively. He received his Ph.D. degree in Computer Science from Dartmouth College, USA, in 2005. He is currently a Professor of Computer Science and Engineering at the University at Buffalo, USA. His research interests include computer vision, machine learning, and media forensics. He is a Fellow of IEEE and IAPR.
\end{IEEEbiography}

\end{document}